\documentclass[runningheads]{llncs}
\usepackage{graphicx}

\usepackage{subfigure}  
\usepackage{color}
\usepackage{amsmath,amssymb,amsfonts}
\usepackage{bm}
\usepackage{multirow}
\usepackage{url}

\begin{document}
\title{EBSD Grain Knowledge Graph Representation Learning for Material Structure-Property Prediction}
\titlerunning{EBSD Grain Knowledge Graph Representation Learning}
%
\author{Chao Shu \and Zhuoran Xin \and Cheng Xie\inst{*}}
\authorrunning{C. Shu et al.}
%
\institute{Yunnan University, School of Software, Kunming 650504, China \\
\email{xiecheng@ynu.edu.cn}}
\maketitle              
\begin{abstract}
The microstructure is an essential part of materials, storing the genes of materials and having a decisive influence on materials' physical and chemical properties. The material genetic engineering program aims to establish the relationship between material composition/process, organization, and performance to realize the reverse design of materials, thereby accelerating the research and development of new materials. However, tissue analysis methods of materials science, such as metallographic analysis, XRD analysis, and EBSD analysis, cannot directly establish a complete quantitative relationship between tissue structure and performance. Therefore, this paper proposes a novel data-knowledge-driven organization representation and performance prediction method to obtain a quantitative structure-performance relationship. First, a knowledge graph based on EBSD is constructed to describe the material's mesoscopic microstructure. Then a graph representation learning network based on graph attention is constructed, and the EBSD organizational knowledge graph is input into the network to obtain graph-level feature embedding. Finally, the graph-level feature embedding is input to a graph feature mapping network to obtain the material's mechanical properties. The experimental results show that our method is superior to traditional machine learning and machine vision methods.

\end{abstract}
\section{Introduction}
 Material science research is a continuous understanding of the organization's evolution, and it is also a process of exploring the quantitative relationship between organizational structure and performance. In the past, the idea of material research was to adjust the composition and process to obtain target materials with ideal microstructure and performance matching. However, this method relies on a lot of experimentation and trial-error experience and is inefficient. Therefore, to speed up the research and development(R\&D) of materials, the Material Genome Project~\cite{jain2013commentary} has been proposed in various countries.
 The idea of the Material Genome Project is to establish the internal connections between ingredients, processes, microstructures, and properties, and then design microstructures that meet the material performance requirements~\cite{jain2013commentary,de2019new}. According to this connection, the composition and process of the material are designed and optimized. Therefore, establishing the quantitative relationship between material composition/process, organizational structure, and performance is the core issue of designing and optimizing materials.

At present, most tissue structure analysis is based on image analysis technology to extract specific geometric forms and optical density data~\cite{Rekha_2017}. However, the data obtained by this method is generally limited to the quantitative information about one-dimensional or two-dimensional images, and it is not easy to directly establish a quantitative relationship between tissue structure and material properties. The method has obvious limitations. In addition, current material microstructure analysis (e.g., metallographic analysis, XRD analysis, EBSD analysis) is often qualitative or partially quantitative and relies on manual experience~\cite {Rekha_2017}. It is still impossible to directly calculate material properties based on the overall organizational structure.

In response to the above problems, this paper proposes a novel data-driven~\cite{zhou2020property} material performance prediction method based on the EBSD~\cite{humphreys2004characterisation}. EBSD is currently one of the most effective material characterization methods. This characterization data not only contains structural information but is also easier for computers to understand. Therefore, we construct a digital knowledge graph~\cite{wang2017knowledge} representation based on EBSD, then design a representation learning network to embed graph features. Finally, we use neural network~\cite{priddy2005artificial} to predict material performance with graph embedding. We conducted experiments on magnesium metal and compared our method with traditional machine learning methods and computer vision methods. The results show the scientific validity of our proposed method and the feasibility of property calculation. The contribution of this page include:
\begin{enumerate}
\item We design an EBSD grain knowledge graph that can digitally represent the mesoscopic structural organization of materials.
\item We propose an EBSD representation learning method that can predict material's performance based on the EBSD organization representation.
\item We establish a database of structural performance calculations that expand the material gene database.
\end{enumerate}

\section{Related work}
\subsection{Data-driven material structure-performance prediction}
Machine learning algorithms can obtain abstract features of data and mine the association rules behind the data. Machine learning algorithms have accelerated the transformation of materials R\&D to the fourth paradigm(i.e., Data-driven R\&D model). Machine learning is applied to material-aided design.

Ruho Kondo et al. used a lightweight VGG16 networks to predict the ionic conductivity in ceramics based on the microstructure picture~\cite{kondo2017microstructure}.
Zhi-Lei Wang et al. developed a new machine learning tool, Material Genome Integrated System Phase and Property Analysis (MIPHA)~\cite{wang2019property}. They use neural networks to predict  the stress-strain curves and mechanical properties based on constructed quantitative structural features.
Pokuri et al. used deep convolutional neural networks to map microstructures to photovoltaic performance, and learn structure-attribute relationships of the data~\cite{pokuri2019interpretable}. They designed a CNN-based model to extract the active layer morphology feature of thin-film OPVs and predict photovoltaic performance.

Machine learning methods based on numerical and visual features can detect the relationship between organization and performance. However, the microstructure of materials contains essential structural information and connection relationships, and learning methods based on descriptors and images will ignore this information.

\subsection{Knowledge graph representation learning}
Knowledge Graph is an important data storage form in artificial intelligence technology. It forms a large amount of information into a form of graph structure close to human reasoning habits and provides a way for machines to understand the world better. Graph representation learning\cite{hamilton2020graph} gradually shows great potential.

In medicine, knowledge graphs are commonly used for embedding representations of drugs. The knowledge graph embedding method is used to learn the embedding representation of nodes directly and construct the relationship between drug entities. The constructed knowledge graphs can be used for downstream prediction tasks. Lin Xuan et al. propose a graph neural network based on knowledge graphs(KGNN) to solve the problem of predicting interactions in drug knowledge graphs~\cite{lin2020kgnn}.

Similarly, in the molecular field, knowledge graphs are used to characterize the structure of molecules/crystals~\cite{wieder2020compact,chen2019graph,jang2020structure}. Nodes can describe atoms, and edges can describe chemical bonds between atoms. The molecular or crystal structure is seen as an individual "graph". By constructing a molecular network map and applying graph representation learning methods, the properties of molecules can be predicted.

In the biological field, graphs are used for the structural characterization of proteins. The Partha Talukdar research group of the Indian Institute of Science did work on the quality assessment of protein models~\cite{2020ProteinGCN}. In this work, they used nodes to represent various non-hydrogen atoms in proteins. Edges connect the K nearest neighbors of each node atom. Edge distance, edge coordinates, and edge attributes are used as edge characteristics. After generating the protein map, they used GCN to learn atomic embedding. Finally, the non-linear network is used to predict the quality scores of atomic embedding and protein embedding.

Compared with the representation of descriptors and visual features, knowledge graphs can represent structural information and related information. The EBSD microstructure of the material contains important grain structure information and connection relationships. Therefore, this paper proposes the representation method of the knowledge graph and uses it for the prediction of organizational performance.

\section{Representation of the EBSD Grain Knowledge Graph}
\begin{figure}[htbp]
\vspace{-1cm}
\centering
\subfigure[Raw scan data]{
\includegraphics[width=5.6cm]{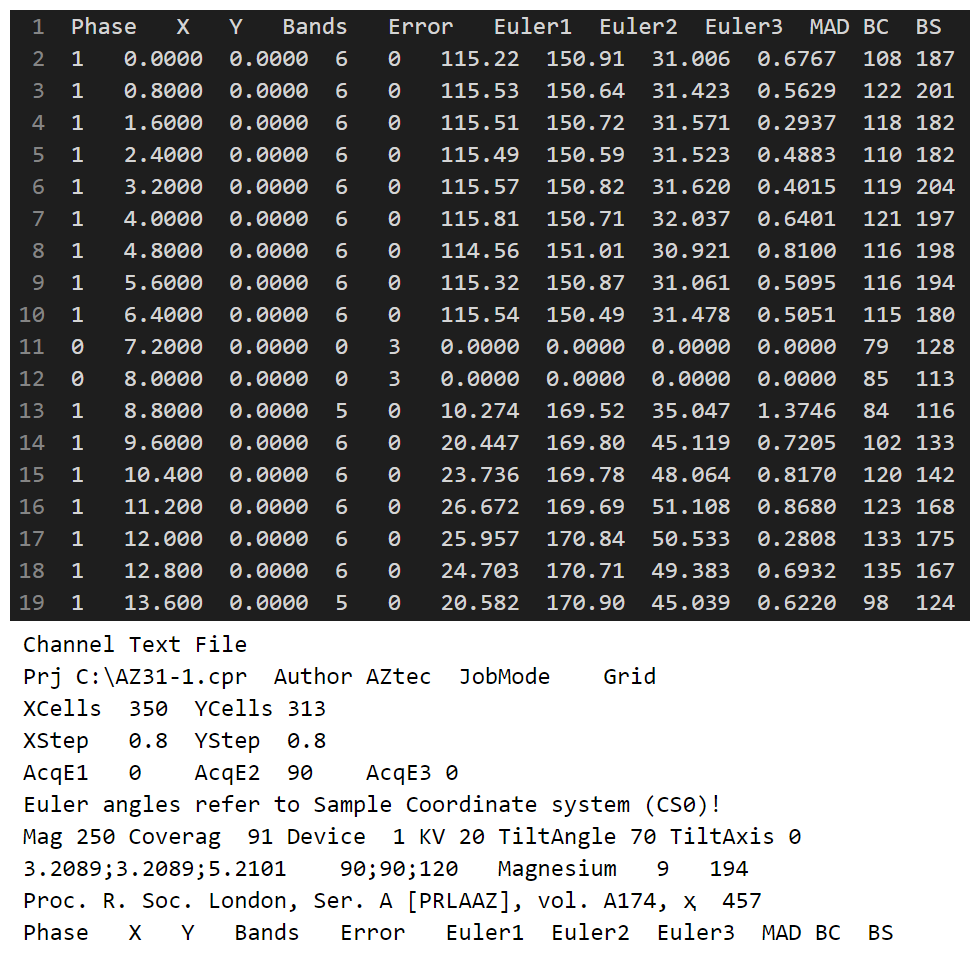}
\label{fig.ebsd.a}
}
\quad
\subfigure[Grain organization map]{
\includegraphics[width=5.4cm]{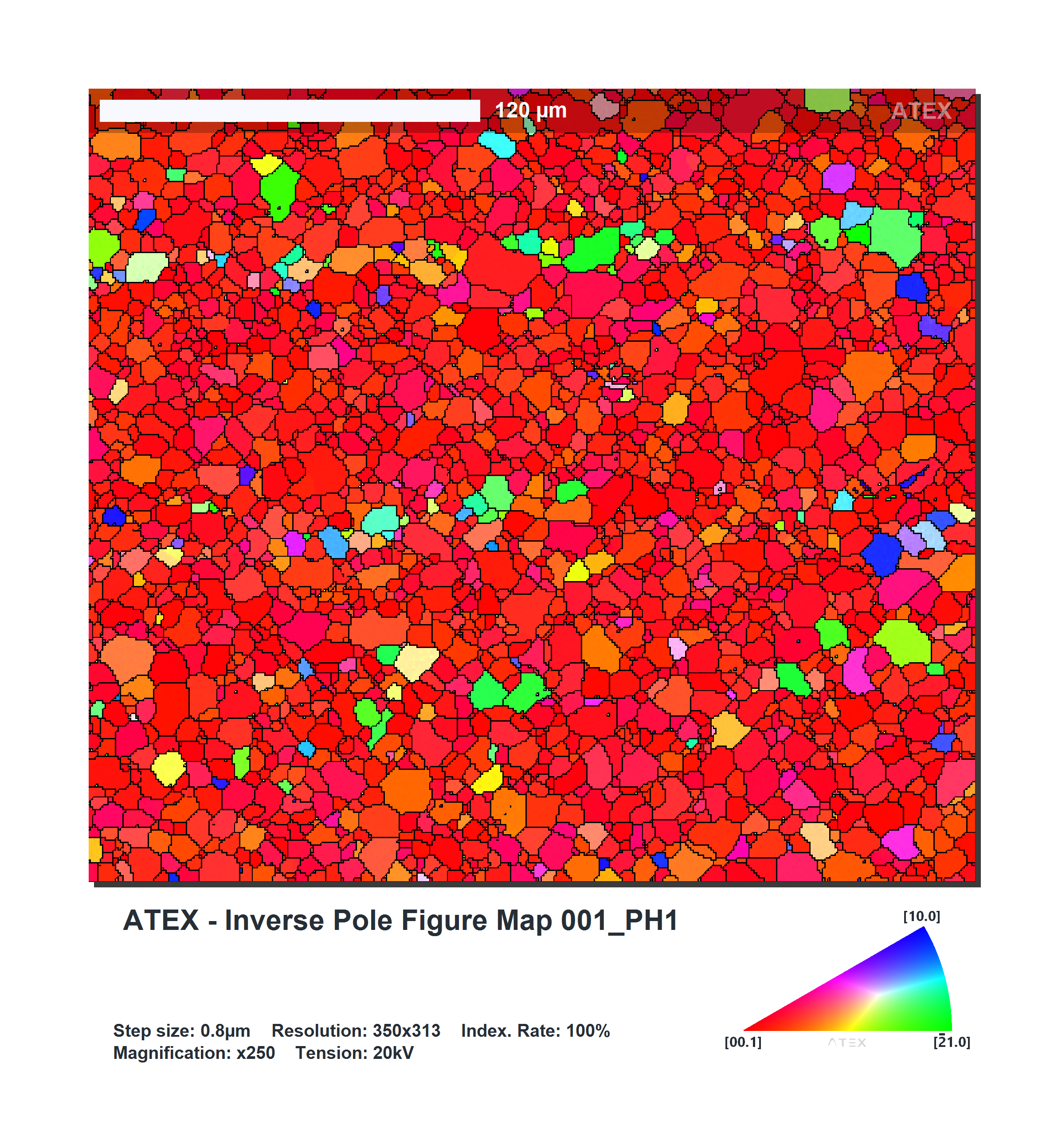}
\label{fig.ebsd.b}
}
\caption{EBSD scan organization information.}
\label{fig.ebsd}
\vspace{-0.5cm}
\end{figure}

In this part of the work, we construct a knowledge graph representation of the micro-organization structure. As shown in the Figure \ref{fig.ebsd}, the left image is the scanning crystallographic data onto the sample, and the right is the Inverse Pole Figure map of the microstructure. The small squares in the Figure \ref{fig.ebsd.b} represents the grains. Based on this grain map data, we construct a grain knowledge graph representation. Because the size, grain boundary, and orientation of the crystal grains affect the macroscopic properties of the material, such as yield strength, tensile strength, melting point, and thermal conductivity~\cite{carneiro2020recent}. Therefore, in this article, we choose the grain as the primary node in the map, and at the same time, we discretize the main common attributes of the grain size and orientation as the attribute node. Then, according to the grain boundaries of the crystal grains, we divided the two adjacent relationships between the crystal grains, namely, strong correlation and weak correlation. Finally, affiliation with grains and attribute nodes is established.

\subsection{Nodes Representation} \label{sec.node_construct}
\subsubsection{Grain node.}
\begin{figure}[htbp]
\includegraphics[width=\textwidth]{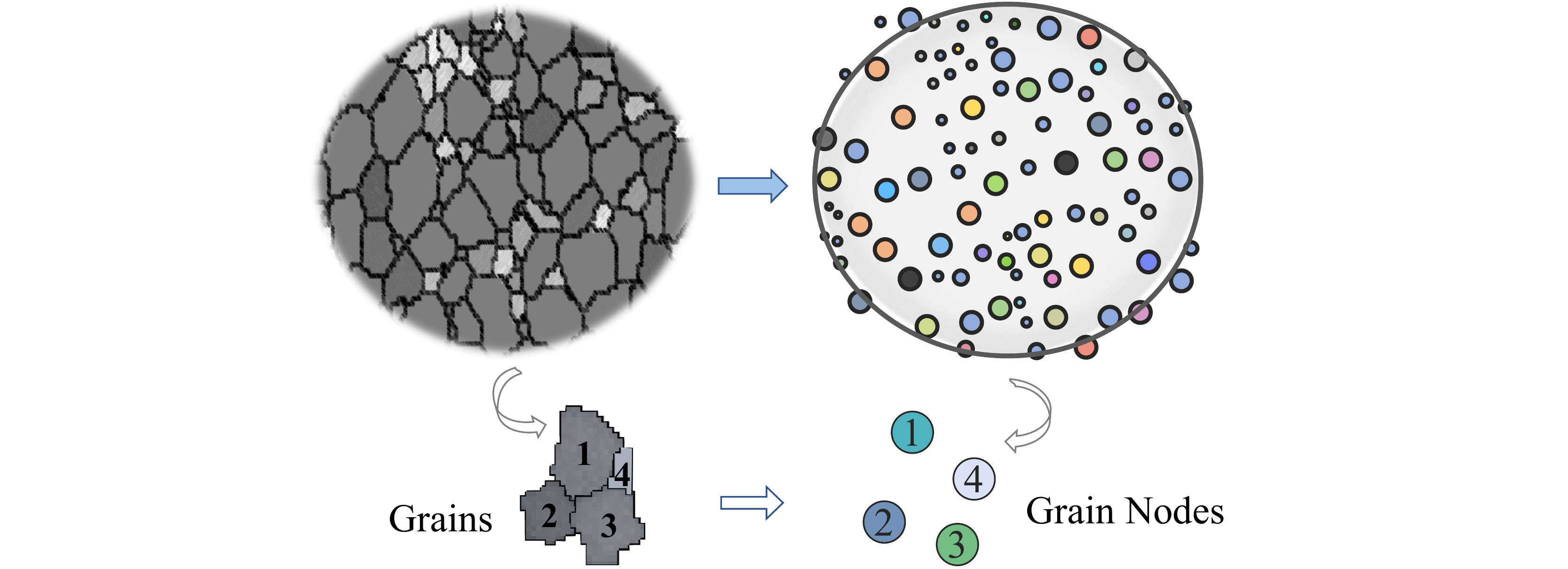}
\caption{The grain node corresponds to the original grain.} \label{fig.grain2node2}
\vspace{-0.5cm}
\end{figure}

We segment each grain in the grain organization map and map it to the knowledge graph as a grain node. First, we use Atex software to count and segment all the grains in a grain organization map.  Then We individually number each grain so that all the grains are uniquely identified, and finally, we build the corresponding nodes in the graph. As shown in the Figure \ref{fig.grain2node2}, the left side corresponds to the grains of the Figure \ref{fig.ebsd.b}, and the right side are the node we want to build. The original grain corresponds to the grain node one-to-one. The grain node is the main node entity in the graph, reflecting the existence and distribution of the grain.

\subsubsection{Grain size attribute node.}
\begin{figure}[htbp]
\vspace{-0.5cm}
\includegraphics[width=\textwidth]{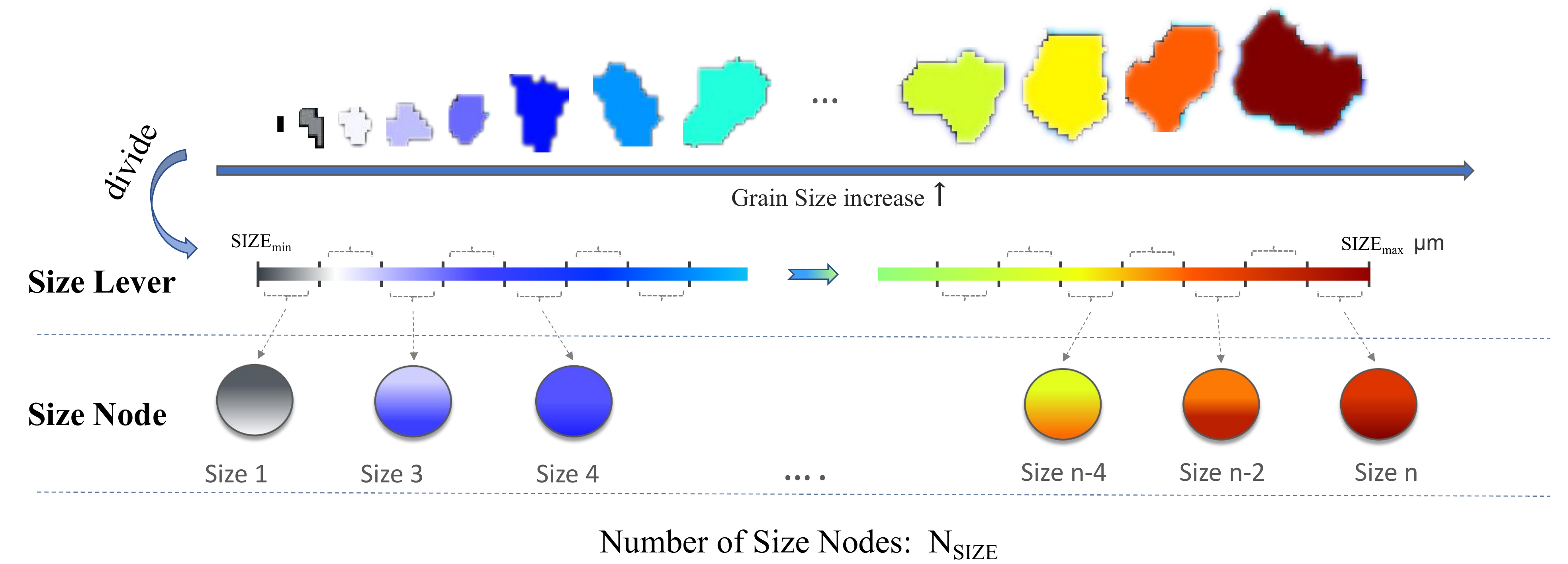}
\caption{Grain size discretization and corresponding size attribute nodes. The size of the grains is marked with different colors. From left to right, the grains are getting bigger and bigger. Then the size interval [$SIZE_{min}$, $SIZE_{max}$] of the grains is found, and this interval is divided into $N_{SIZE}$ parts. Finally, a size node is constructed for each divided interval.} \label{fig.size2node}
\vspace{-1cm}
\end{figure}

Next, we construct grain size attribute nodes used to discretize and identify the grain size. First, we discretize the size of the crystal grains. As shown in the Figure \ref{fig.size2node}, the color represents the difference in the size of the grains, $SIZE_{max}$ represents the largest-scale grain size, and $SIZE_{min}$ represents the smallest-scale grain size. The grain size levels are divided into $N_{SIZE}$, and the interval size of each level is $(SIZE_{max}-SIZ_{min})/N_{SIZE}$. We regard each interval as a category, as shown in the equation \ref{eq1}, for each grain, we divide it into corresponding category according to its size. We use the discretization category to represent the grain size instead of the original value. Then we construct a size attribute node for each size category, as shown in Figure \ref{fig.size2node}. Finally, we use the one-hot method to encode these $N_{SIZE}$ categories and use the one-hot encoding as the feature of the size attribute node.
\begin{equation}
L\_S_{node}=\lceil Grain.size/\lceil (SIZE_{max}-SIZE_{min})/N_{SIZE}\rceil\rceil
\label{eq1}
\end{equation}
where $L\_S_{node}$ represents the size category of the grain. $Grain.size$ is the circle equivalent diameter of the grain, $\lceil\rceil$ means rounding up.

\subsubsection{Grain orientation attribute node.}
\begin{figure}[htbp]
\vspace{-0.5cm}
\includegraphics[width=\textwidth]{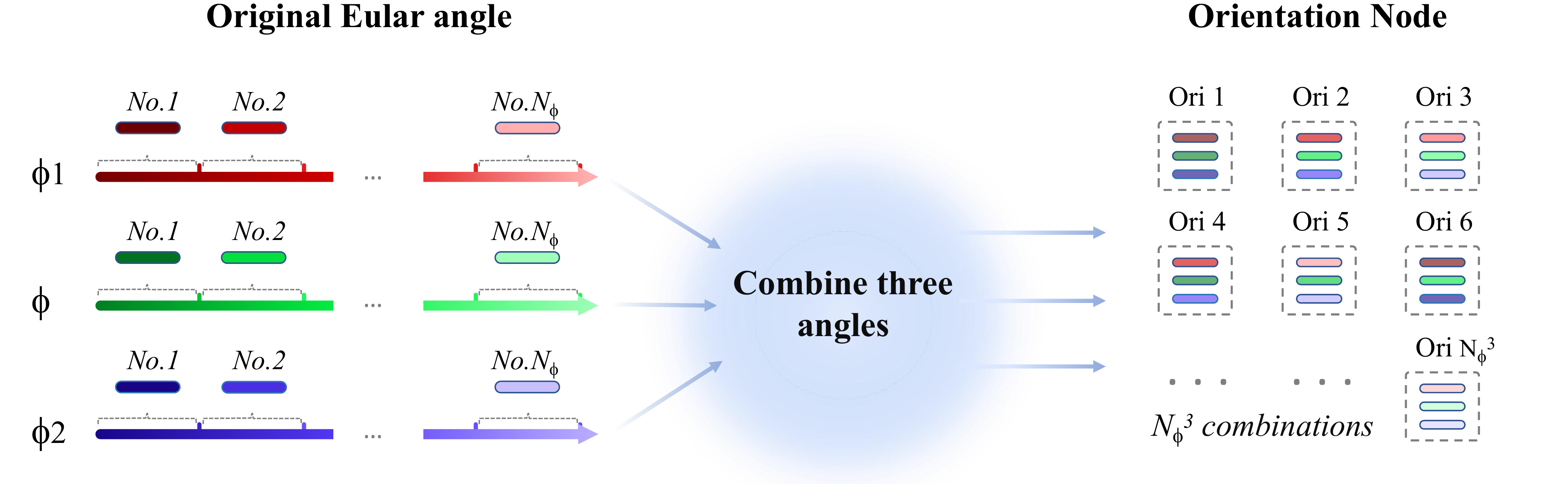}
\caption{Grain orientation discretization and corresponding orientation attribute nodes. On the left are three directions Euler angles, whose angles are represented by the RGB color. Each Euler angle is divided into $N_{\phi}$ parts, and then each equal part of each Euler angle is combined with one equal part of the remaining Euler angle. Each combination is regarded as an orientation category, and a node is constructed for this.}
\label{fig.ori2node}
\vspace{-0.4cm}
\end{figure}

In this work, Euler angles are used to identify the orientation of grains. Similarly, we also discretize the Euler angles, as shown in Figure \ref{fig.ori2node}. The orientation of the grains is determined by the euler angles in three directions, so we discretize the euler angles in the three directions and combine them. The obtained three Euler angle interval combinations are the discretized types of orientation. Specifically as shown in the equation \ref{eq2}, we first calculate the maximum and minimum values of the three Euler angles $\phi(\phi1, \phi, \phi2)$ for all grains, namely $\bm{ \phi_{max}}=\{\phi1_{max}, \phi_{max}, \phi2_{max}\}, \bm{\phi_{min}}=\{\phi1_{min}, \phi_{ min}, \phi2_{min}\}$. Then each Euler angle $\phi(\phi1, \phi, \phi2)$ is divided into $N_\phi$ equal parts, the
length of each part is $( \bm{\phi_{max}}-\bm{\phi_{min}})/N_\phi$. Finally, the $N_\phi$ equal parts of each Euler angle are cross-combined to obtain $N_ \phi^3$ combinations. We regard each combination as a kind of orientation, i.e., there are $N_\phi^3$ orientation categories. For each grain, we can map it to one of $N_\phi^3$ categories according to its three Euler angles $\phi(\phi1, \phi, \phi2)$, As shown in the equation \ref{eq2}. In this way, all crystal grains are divided into a certain type of orientation. We construct an orientation attribute node for each type of orientation to represent orientation information. Similarly, we use the one-hot method to encode these $N_\phi^3$ categories individually. Each orientation category will be represented by a $N_\phi^3$-dimensional one-hot vector used as the feature of the corresponding orientation attribute node.
\begin{equation}
\begin{aligned}
L\_O_{node} =  \{
&\lceil Grain.\phi1/ \lceil(\phi1_{max}-\phi1_{min})/N_\phi \rceil \rceil, \\
&\lceil Grain.\phi/ \lceil(\phi_{max}-\phi_{min})/N_\phi \rceil \rceil,   \\
&\lceil Grain.\phi2/ \lceil(\phi2_{max}-\phi2_{min})/N_\phi \rceil \rceil\}
\end{aligned}
\label{eq2}
\end{equation}
where $Grain.\phi1$, $Grain.\phi$ and $Grain.\phi2$ are the euler angles in the three directions. $L\_O_{node}$ represents the orientation category to which the grains are classified. $\lceil\rceil$ refers to rounding up.

\subsection{Edges Representation}
After the construction of the node, the edges between the nodes need to be constructed. The nodes reflect the entities in the graph, and the edges contain the structural information of the graph. We build edges in the grain knowledge graph based on crystallographic knowledge. The constructed edge represents the association between nodes, including position association and property association. The edges between grain nodes reflect position information and grain boundaries; the edges between grain nodes and grain attribute nodes describe the properties of the grains.

\subsubsection{Edge between grain nodes.}
\begin{figure}[htbp]
\vspace{-0.5cm}
\centerline{\includegraphics[width=\textwidth]{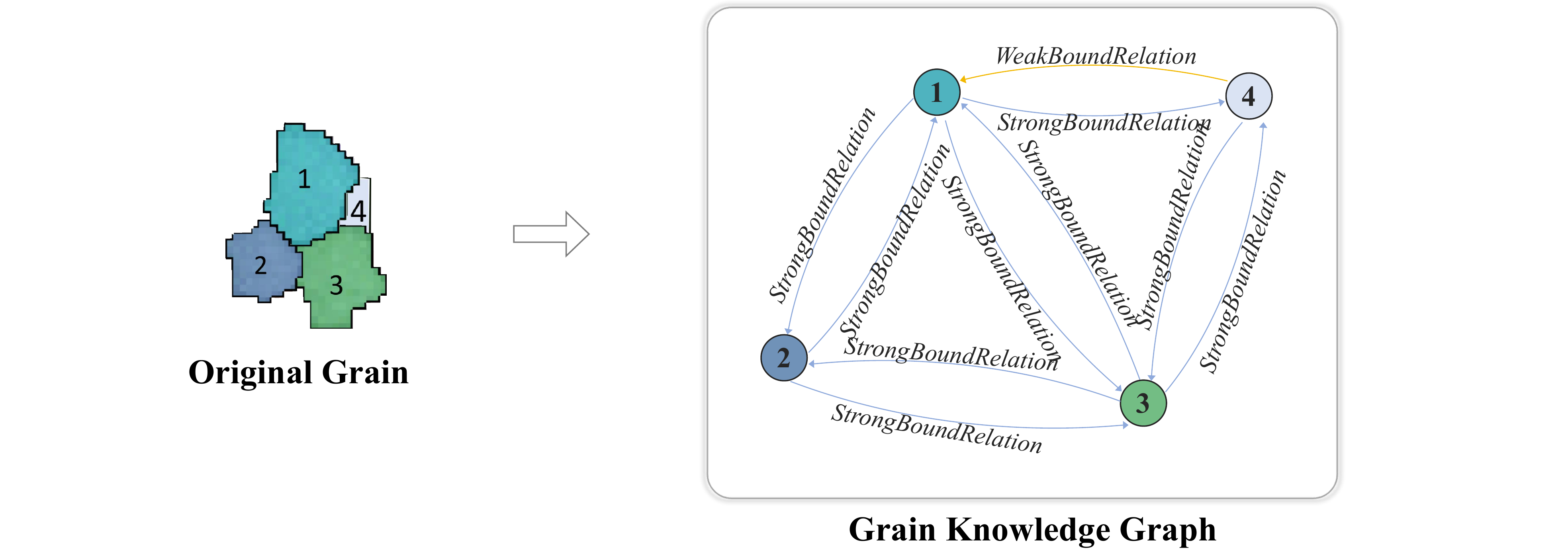}}
\caption{Adjacent grains and the edges between them}
\label{fig.ref_G2G}
\vspace{-0.65cm}
\end{figure}

The contact interface between the grains is called the grain boundary, representing the transition of the atomic arrangement from one orientation to another. Generally speaking, grain boundaries have a significant impact on the various properties of the metal. In order to describe the boundary information of grains, we construct edges between grain nodes. First, we obtain the neighboring grains of each grain and construct the connection between the neighbor grain nodes. In order to further restore more complex grain spatial relationships, we set up a knowledge of neighboring rules. As shown in the equation \ref{eq3}, we use $lp$ to represent the ratio of the bordering edge length of the grain to the total perimeter of the grain. Then we set a threshold $\lambda$, as shown in the equation \ref{eq4}, when the $lp$ of grain A and grain B is greater than or equal to $\lambda$, we set the relationship between the A node and the B node to be a strong correlation; otherwise, it is set to weak correlation. Figure \ref{fig.ref_G2G} shows the edge between grain nodes.
\begin{equation}
lp = bound\_length / perimeter
\label{eq3}
\end{equation}
\begin{equation}
Rel\_G\_G(lp)=\begin{cases}
\text{Strong association}, & lp < \lambda \\
\text{Weak association},   & lp \ge \lambda
\end{cases}
\label{eq4}
\end{equation}

\subsubsection{Edge between grain node and size attribute node.}
\begin{figure}[htbp]
\vspace{-0.5cm}
\centerline{\includegraphics[width=\textwidth]{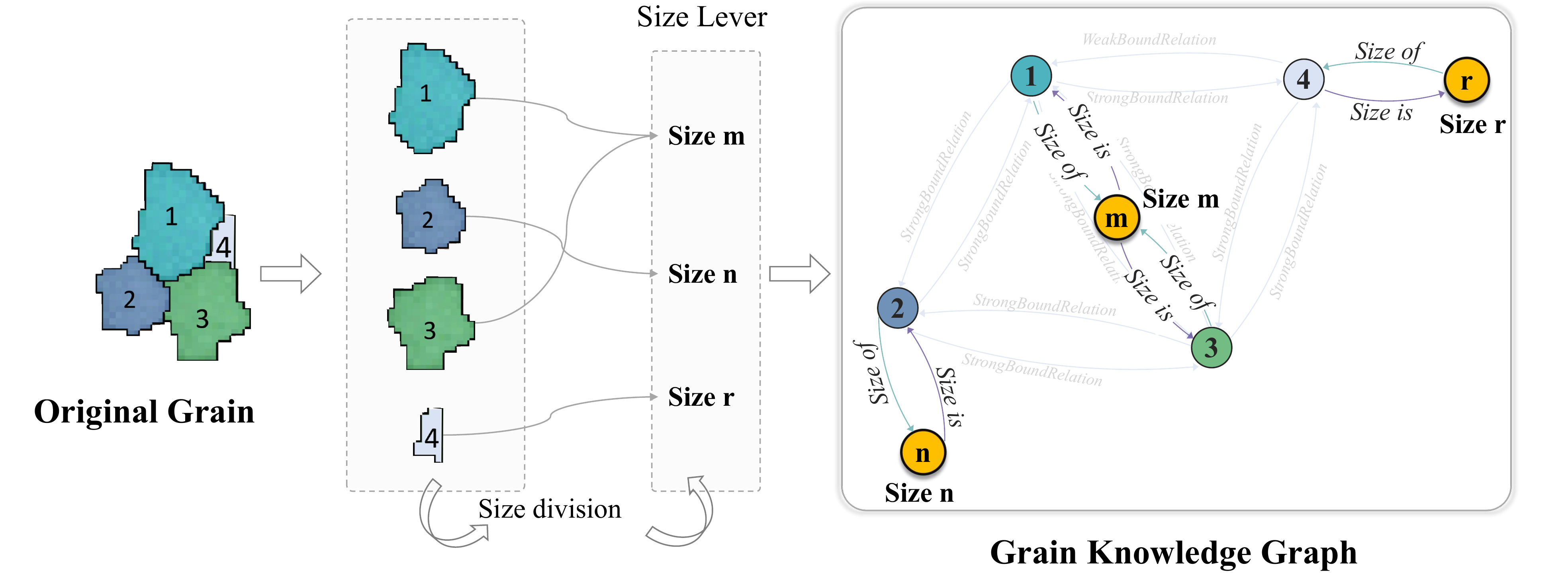}}
\caption{Edge between grain node and size attribute node.}
\label{fig.ref_G2S}
\vspace{-0.65cm}
\end{figure}

In section \ref{sec.node_construct}, we construct two types of attribute nodes. Here we associate the grain node with the attribute node to identify the property of the grain. First, we calculate the grain size category according to the equation \ref{eq1}, and then associate the corresponding grain node with the corresponding size attribute node to form the edge. As shown in the Figure \ref{fig.ref_G2S}, we calculate the size categories $\lbrace$m, m, n, r$\rbrace$ of the four grains $\lbrace$1, 2, 3, 4$\rbrace$, and then associate the corresponding grain node with the size attribute nodes to form the belonging relationship.

\subsubsection{Edge between grain node and orientation attribute node.}
\begin{figure}[htbp]
\vspace{-0.85cm}
\centerline{\includegraphics[width=\textwidth]{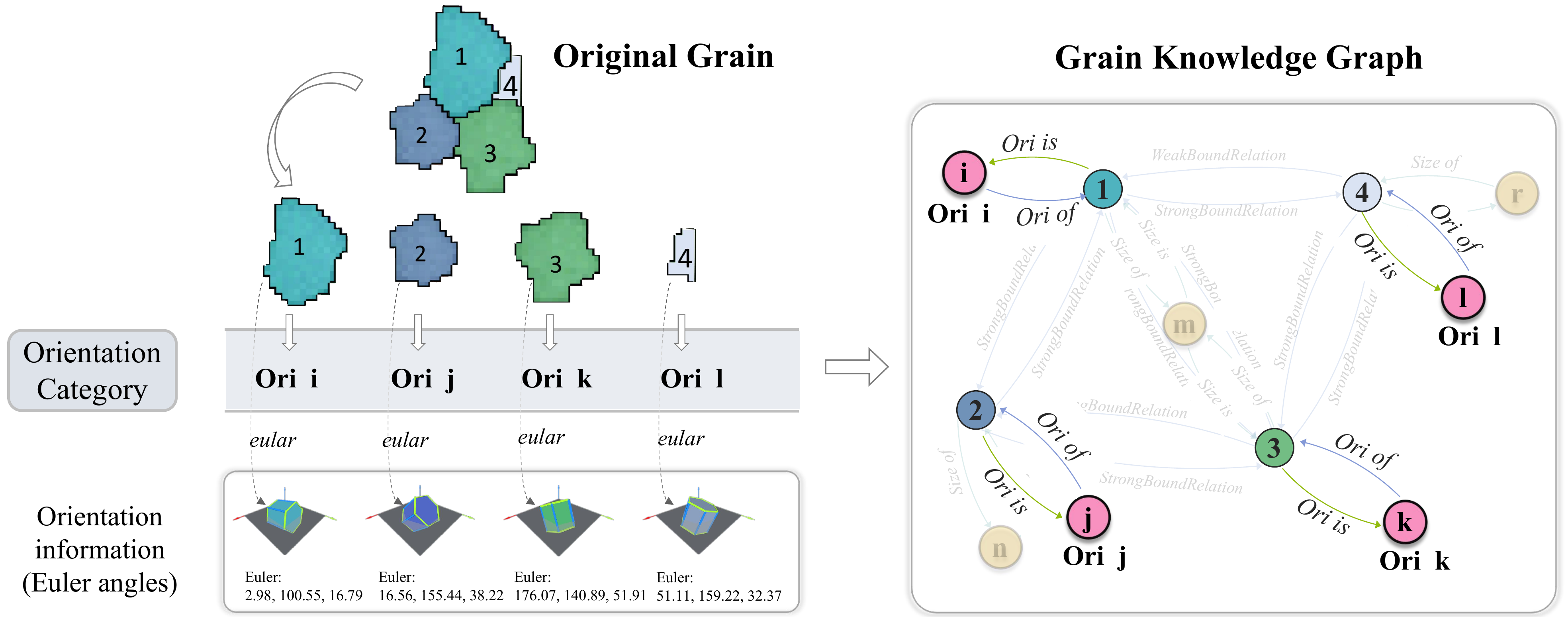}}
\caption{Edge between grain node and orientation attribute node.}
\label{fig.ref_G2O}
\vspace{-0.75cm}
\end{figure}

Similarly, we identify the orientation category for the grain node by associating it with the orientation attribute node. As shown in the Figure \ref{fig.ref_G2O}, first we split the grains in the map, the bottom left of the picture shows the three Euler angles of the grains. Then we calculate the orientation category $\lbrace$i, j, k, l$\rbrace$ of the grains $\lbrace$1, 2, 3, 4$\rbrace$ according to equation \ref{eq3}. Finally, we associate the corresponding orientation attribute node with the corresponding grain node. Figure \ref{fig.ref_G2O} shows the edges between the grain nodes and the orientation attribute nodes, reflecting the discrete orientation characteristics and orientation distribution of the grains.

\subsection{Grain graph convolutional prediction model}
\begin{figure}[htbp]
\vspace{-0.3cm}
\centerline{\includegraphics[width=\textwidth]{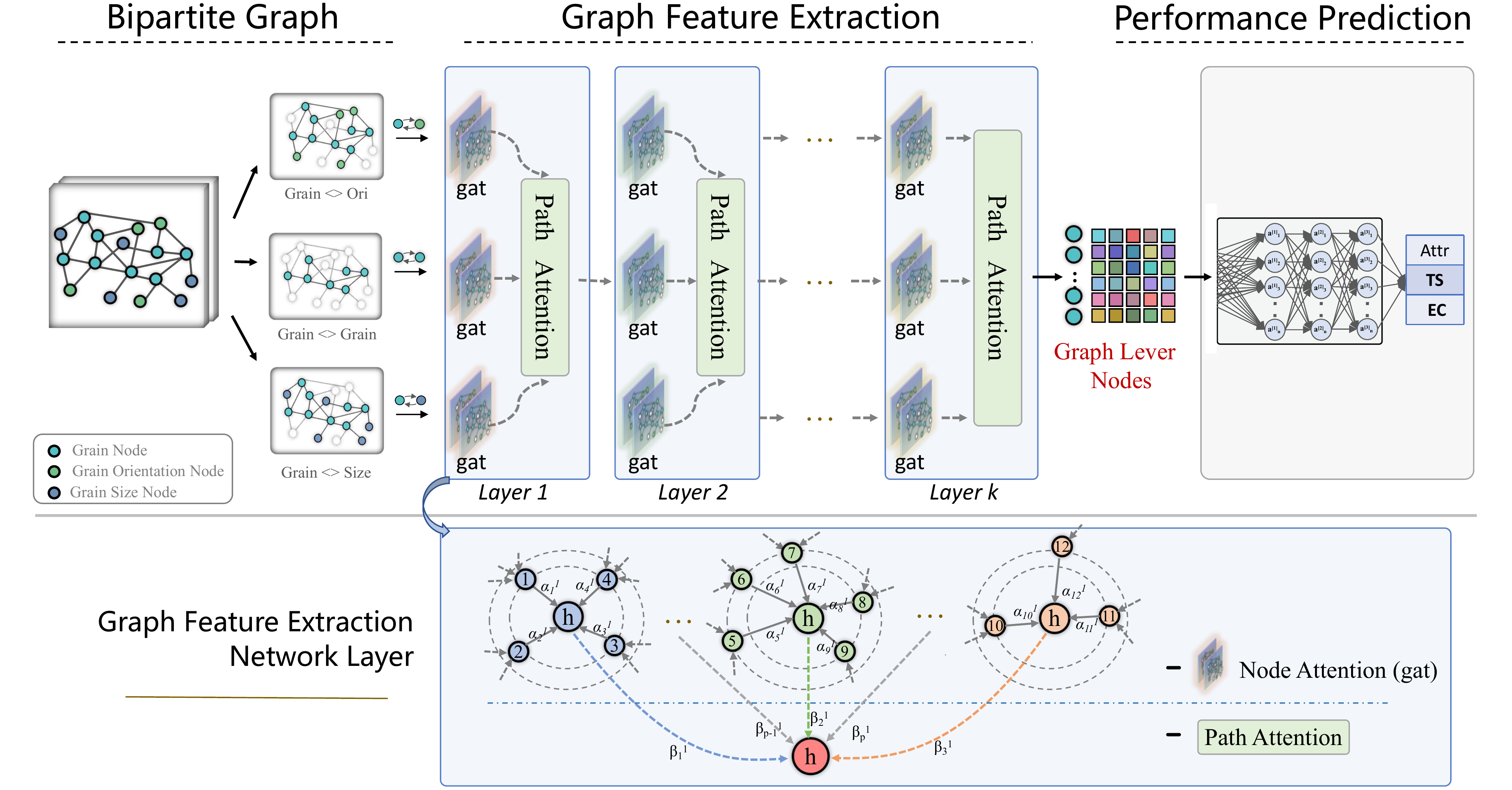}}
\caption{Grain graph convolutional prediction model. The model is divided into graph feature extraction part and performance prediction part. The graph feature extraction network is composed of multiple node-level graph attention networks(gat) and a path-level attention aggregation network. The prediction network is a multilayer neural network. The graph feature network extracts graph-level features, and the prediction network maps graph-level features to material properties.}
\label{fig.model}
\vspace{-0.5cm}
\end{figure}

The structured grain knowledge graph can describe the microstructure of the material. Next, we build a graph feature convolution network(grain graph convolutional network) to embed the grain knowledge graph and realize graph feature extraction. Then, a feature mapping network based on a neural network is built to predict material properties with the graph feature. The complete model we built is shown in Figure \ref{fig.model}.

\subsubsection{Grain graph convolutional network.}
The graph features convolution network is a heterogeneous graph convolution network~\cite{wang2019heterogeneous}. First, the heterogeneous grain knowledge graph is divided into multiple bipartite graphs and isomorphic graphs according to the type of edges. Next, the features of the nodes in the meta-path of subgraph are transferred and aggregated. Then the features of the same nodes of the subgraphs are fused, and finally, the graph-level characterization nodes are obtained through multiple convolutions. The process of graph convolution is shown at the bottom of Figure \ref{fig.model}. Specifically, the node aggregation process includes node-level feature aggregation and path-level feature aggregation. In node message transmission, we use node-level attention to learn the attention value of adjacent nodes on the meta-path. After completing the message transmission of all meta-paths, we use path-level attention to learn the attention value of the same nodes on different meta-paths. With the double-layer attention, the model can capture the influence factors of nodes and obtain the optimal combination of multiple meta-paths. Moreover, the nodes in the graph can better learn complex heterogeneous graphical and rich information. The equation \ref{eq5} shows the feature aggregation transformation under the node-level attention. $\iota$ represents different paths/edges, and there are a total of $p$ edges. $\alpha_{ij}$ represents the attention score between node i and node j. $LeakyReLU$ and $Softmax$ are activation functions, $W$ is a learnable weight matrix, and $\vec{a}$ is a learnable weight vector. $\parallel$ represents concatenation, and $N(i)$ refers to all neighbor nodes of node i. $h_i^{k+1}$ represents the $(k+1)$ layers embedding of node i.
\begin{equation}
\begin{split}
\vec{z}_i^{k^{(\iota)}}     &=  \bm{W}^{k^{(\iota)}} \cdot \vec{h}_i^{k^{(\iota)}} \\
     e_{ij}^{k^{(\iota)}}   &=  LeakyReLU(\vec{a}^{k^{(\iota)}} \cdot [\vec{z}_i^{k^{(\iota)}} \parallel \vec{z}_j^{k^{(\iota)}}]) \\
\alpha_{ij}^{k^{(\iota)}}   &=  Softmax_j(e_{ij}^{k^{(\iota)}}) \\
\vec{h}_i^{{k+1}^{(\iota)}} &=  \sigma(\sum\limits_{j \in {N(i)}^{(\iota)}} \alpha_{ij}^{k^{(\iota)}} \cdot \vec{z}_i^{k^{(\iota)}})
\end{split}
\label{eq5}
\end{equation}

The equation \ref{eq6} shows the change of node characteristics at the path level, $\beta_{(\iota)}^k$ is the important coefficient of each meta-path.  We first perform a nonlinear transformation on the output $\vec{h}_i^{{k+1}^{(\iota)}}$ of the node-level attention network, and then perform a similarity measurement with a learnable attention vector $q$. Next, we input the result of the similarity measurement into the Softmax function to obtain important coefficients, and finally perform weighted summation on the node embeddings on each meta-path. After completing multiple graph feature convolutions, we obtain graph-level node embeddings.
\begin{equation}
\begin{split}
\beta_{(\iota)}^k  &=  Softmax(\frac{1}{N(i)} \sum\limits_{\iota \in N(i)} \vec{q} \cdot tanh(\bm{W}^k \cdot \vec{h}_i^{{k+1}^{(\iota)}} + \vec{b}) ) \\
  \vec{h}_i^{k+1}  &=  \sum\limits_{\iota = 1}^p \beta_{(\iota)}^k \cdot \vec{h}_i^{{k+1}^{(\iota)}}
\end{split}
\label{eq6}
\end{equation}

\subsubsection{Feature mapping network.}The microstructure-performance relationship is usually qualitatively studied through statistical methods (e.g., statistics of grain size, orientation, and grain boundaries). The relationship between the microstructure and properties is difficult to obtain through comparative observation or direct calculation. However, Artificial neural networks can mine more essential characteristics of data and establish complex relationships between data~\cite{priddy2005artificial}. Here, we have used the graph features convolution network to extract the features of the grain knowledge graph, so we use a feature mapping network based on a neural network to implement machine learning tasks. As shown in equation \ref{eq7}, $\vec{h}_i$ is the final graph-level node vector, $fc$ is the mapping network. The network comprises a data normalization layer, a fully connected layer, an activation layer, and a random deactivation layer. Through this network, the feature of the grain knowledge graph can be mapped to the property of material.
\begin{equation}
prop = \frac{1}{n} \sum_{i=1}^{n} fc(\vec{h}_i)
\label{eq7}
\end{equation}

\section{Experiment}
\subsection{Dataset}
The experimental data comes from the EBSD experimental data of 19 Mg metals and also includes the yield strength(ys), tensile strength(ts), and elongation(el) of the sample. The EBSD scan data contains a total of 4.46 million scan points. As a result, the number of nodes in all the constructed knowledge graphs is 40,265, and the number of edges reaches 389,210. We use EBSD knowledge graph representation as model input and mechanical properties as label.

\subsection{Comparison methods and Results}
We design two different methods to compare with ours. They are traditional machine learning methods based on statistics, image feature extraction methods based on computer vision. We use traditional machine learning methods to directly calculate the attribute characteristics of all grains to obtain material properties. These methods include Ridge, SVR, KNN, ExtraTree. 
In addition, we use the pre-trained CNN model to directly learn visual features from the microstructure map and predict performance. The model is Resnet-50.

The model performance evaluation results are shown in Table \ref{tb2}. It can be seen that our method is superior to other methods. Our method obtained an R2 value of 0.74. This shows that our method can extract more effective features. Traditional machine learning methods and machine vision methods have obtained acceptable R2 values, which shows that both methods can obtain microstructure characteristics to a certain extent. However, compared with traditional machine learning, the method of machine vision does not show much superiority. It is because CNN training requires a larger amount of data, and our current data set is small.

\begin{table}
\vspace{-0.4cm}
\centering
\caption{Results of model for ys prediction}
\label{tb2}
\setlength{\tabcolsep}{2.74mm}{
\begin{tabular}{llll}
\hline
Model
              &MSE    &MAE   &R2      \\  \hline
Ridge         &112.5  & 6.9  &0.590   \\
SVR           &102.7  & 4.8  &0.626   \\
KNN           & 97.6  & 3.6  &0.651   \\
ExtraTree     &105.9  & 5.6  &0.610   \\
\hline
Resnet50      & 94.8  & \textbf{3.1}  &0.667   \\
\hline
\textbf{Hetero\_GAT(Our)}
            &\textbf{73.1}   &5.9    &\textbf{0.74}  \\ \hline
\end{tabular}}
\vspace{-0.5cm}
\end{table}

\section{Conclusion}
This paper proposes a novel material organization representation and performance calculation method. First, we use the knowledge graph to construct the EBSD representation. Then, we designed a representation learning network to abstract the EBSD representation as graph-level features. Finally, we built a neural network prediction model to predict the corresponding attributes. The experimental results prove the effectiveness of our method. Compared with traditional machine learning methods and machine vision methods, our method is more reasonable and practical.


%
%
%

\bibliographystyle{splncs04}
\bibliography{mybibliography}

\end{document}